\documentclass{article}

\usepackage[preprint]{neurips_2026}

\usepackage[utf8]{inputenc} 
\usepackage[T1]{fontenc}    
\usepackage{hyperref}       
\usepackage{url}            
\usepackage{booktabs}       
\usepackage{amsfonts}       
\usepackage{nicefrac}       
\usepackage{microtype}      
\usepackage{xcolor}         

\usepackage{amsmath} 
\usepackage{booktabs}
\usepackage{multirow}
\usepackage[table]{xcolor}
\usepackage{graphicx}
\usepackage{wrapfig}

\title{VEGA: Visual Encoder Grounding Alignment for Spatially-Aware Vision-Language-Action Models}

\author{
  \textbf{Hao Wang$^{1}$, Xiaobao Wei$^{1,2}$, Jingyang He$^{1,2}$, Chengyu Bai$^{1,2}$,} \\ 
  \textbf{Chun-Kai Fan$^{1,2}$, Jiajun Cao$^{1,2}$, Jintao Chen$^{1,2}$, Ying Li$^{1}$, Shanyu Rong$^{1}$} \\
  \textbf{Ming Lu$^{1}$, Xiaozhu Ju$^{1,2}$, Jian Tang$^{1,2}$, Shanghang Zhang$^{1}$\thanks{Corresponding author}} \\
  $^{1}$Peking University, Beijing, China \\ 
  $^{2}$Beijing Innovation Center of Humanoid Robotics, Beijing, China \\
}

\begin{document}

\maketitle

\begin{abstract}
Precise spatial reasoning is fundamental to robotic manipulation, yet the visual backbones of current vision-language-action (VLA) models are predominantly pretrained on 2D image data without explicit 3D geometric supervision, resulting in representations that lack accurate spatial awareness. Existing implicit spatial grounding methods partially address this by aligning VLA features with those of 3D-aware foundation models, but they rely on empirical layer search and perform alignment on LLM-level visual tokens where spatial structure has already been entangled with linguistic semantics, limiting both generalizability and geometric interpretability. We propose VEGA (Visual Encoder Grounding Alignment), a simple yet effective framework that directly aligns the output of the VLA's visual encoder with spatially-aware features from DINOv2-FiT3D, a DINOv2 model fine-tuned with multi-view consistent 3D Gaussian Splatting supervision. By performing alignment at the visual encoder output level, VEGA grounds spatial awareness before any linguistic entanglement occurs, offering a more interpretable and principled alignment target. The alignment is implemented via a lightweight projector trained with a cosine similarity loss alongside the standard action prediction objective, and is discarded at inference time, introducing no additional computational overhead. Extensive experiments on simulation benchmark and real-world manipulation tasks demonstrate that VEGA consistently outperforms existing implicit spatial grounding baselines, establishing a new state-of-the-art among implicit spatial grounding methods for VLA models.
\end{abstract}

\section{Introduction}

Robotic manipulation requires precise spatial reasoning about objects, contacts, and intricate geometric relationships in the 3D physical world~\citep{song2025robospatial,landsiedel2017review}. Recent vision-language-action (VLA) models~\citep{kim2024openvla, kim2025fine, black2024pi_0, black2025pi_, brohan2022rt, zitkovich2023rt, bjorck2025gr00t} have demonstrated remarkable instruction-following capabilities by leveraging large-scale vision-language pretraining, yet their visual backbones are predominantly trained on unstructured 2D image data without access to any explicit 3D geometric supervision~\citep{liu2023visual, chen2023pali, karamcheti2024prismatic}. As a consequence, the visual representations learned by these models lack accurate spatial awareness, limiting their ability to reason about object depths, spatial configurations, and viewpoint variations that are essential for precise manipulation in the real world~\citep{chen2024spatialvlm, kamath2023s}. While such models can succeed on tasks where coarse 2D appearance cues are sufficient, they tend to generalize poorly when manipulation depends on fine-grained spatial understanding, such as estimating object height, inferring relative positions, or adapting to novel viewpoints~\citep{shridhar2023perceiver, goyal2023rvt, yuan2024robopoint}. 

\begin{figure}[!t]
  \centering
  \includegraphics[width=\textwidth]{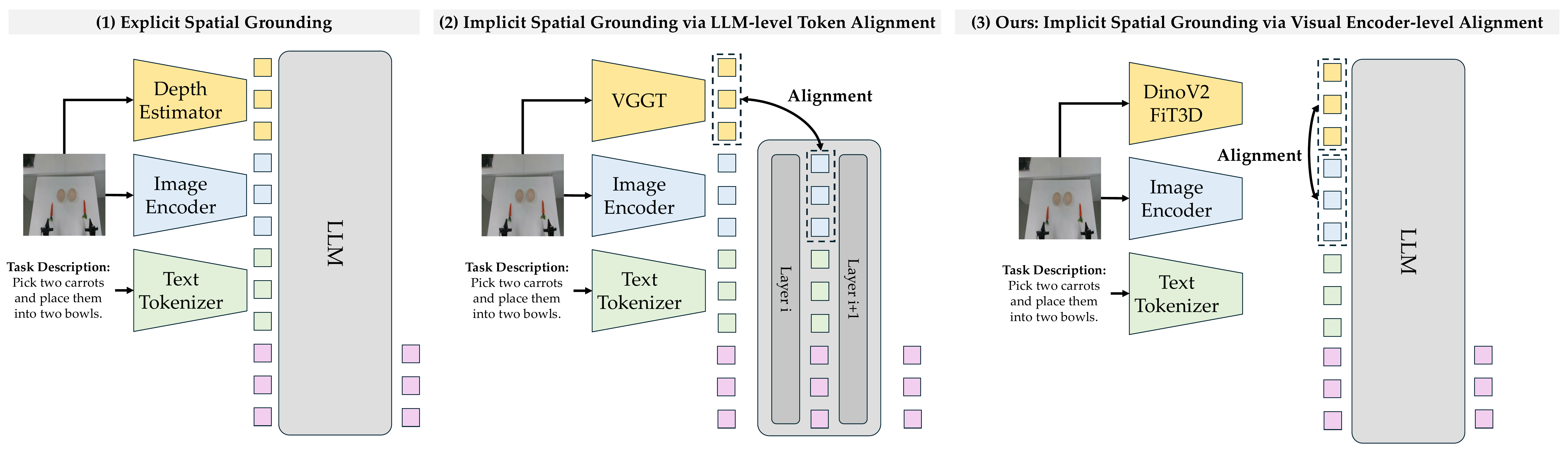}
  \caption{\textbf{Comparison of spatial grounding paradigms for VLA models.} (1) \textit{Explicit Spatial Grounding} augments VLA inputs with estimated depth maps from monocular depth estimators (e.g., Depth Anything), introducing additional inference overhead and error propagation. (2) \textit{Implicit Spatial Grounding via LLM-level Token Alignment} aligns visual features at the LLM token level, where spatial structure is entangled with linguistic semantics. (3) \textit{Ours (VEGA)} directly aligns visual encoder outputs with spatially-aware features via a lightweight projector, grounding spatial awareness before any linguistic entanglement occurs.}
  \label{fig:teaser}
  \vspace{-1em}
\end{figure}

To bridge this gap, existing approaches generally fall into two categories.
The first category explicitly incorporates 3D information into VLA models, either by augmenting inputs with depth maps or point clouds from additional sensors~\cite{qu2025spatialvla, sun2025geovla, bhat20253d}, or by estimating 3D structure from 2D images via depth estimators~\cite{yang2024depth, yang2024depthv2}.
Additionally, some approaches directly supervise the VLA's geometric reasoning by fine-tuning on curated robotic datasets enriched with human-annotated 3D bounding boxes, trajectory keypoints, or spatial coordinates~\citep{zhen20243d, huang2023embodied}.
While effective, these methods face practical limitations: depth sensors introduce hardware and calibration overhead, depth maps are often noisy for transparent or reflective objects~\citep{kroemer2021review}, and large-scale 3D annotated datasets remain scarce and costly to collect~\citep{o2024open, deitke2023objaverse}.
The second category pursues a more hardware-agnostic alternative via implicit spatial grounding, incorporating features from frozen 3D foundation models~\citep{wang2025vggt} into VLA representations through representation alignment supervision~\citep{li2025spatial, sun2026rocket, guo2025glad} or feature fusion~\citep{lin2025evo}. 
This direction is appealing precisely because it requires no additional sensors or 3D annotations at inference time. 
However, these methods share two fundamental limitations.
First, they rely on empirical layer search to identify which intermediate layer to align, making the alignment target sensitive to hyperparameter choice and difficult to generalize. Second, alignment is performed on LLM-level visual tokens where visual features have already been fused with linguistic context, conflating geometric structure with semantic associations and losing geometric interpretability.

To address this limitation, we propose \textbf{VEGA} (\textbf{V}isual \textbf{E}ncoder \textbf{G}rounding \textbf{A}lignment), a simple yet effective representation alignment framework that directly aligns VLA visual encoder outputs with spatially-aware features from a 3D-aware foundation model. 
Rather than relying on empirical layer search or aligning LLM-level tokens, VEGA supervises the visual encoder outputs of the VLA backbone with features extracted from FiT3D~\citep{yue2024improving}, a DINOv2 model fine-tuned with 3D-aware supervision via multi-view consistent 3D Gaussian Splatting~\citep{kerbl20233d}.
By performing alignment directly at the visual encoder level, VEGA ensures that spatial awareness is grounded at the representation stage before linguistic entanglement occurs, offering a more geometrically interpretable alignment target. This design introduces no additional computational overhead at inference time. 
Extensive experiments on the RoboTwin 2.0 benchmark~\citep{chen2025robotwin} and real-world robotic manipulation tasks demonstrate that VEGA consistently outperforms existing methods, establishing a new state-of-the-art among implicit spatial grounding methods for VLA models.

In summary, our main contributions are as follows:
\begin{itemize}
\item We identify that existing implicit spatial grounding methods suffer from two compounding issues: reliance on empirical layer search, and performing alignment on LLM-level tokens where geometric structure has been conflated with linguistic semantics. We show that directly aligning at the visual encoder output level resolves both issues simultaneously.
\item We propose VEGA, a simple yet effective alignment framework that supervises VLA visual encoder outputs with DINOv2-FiT3D features via a lightweight projector, injecting 3D spatial awareness at the most geometrically meaningful representation stage without any additional inference overhead.
\item Extensive experiments on RoboTwin and real-world manipulation tasks demonstrate that VEGA consistently outperforms existing methods, achieving state-of-the-art performance among implicit spatial grounding methods for VLA models.
\end{itemize}

\section{Related Work}

\subsection{Vision-Language-Action Models}

The integration of Large Language Models (LLMs) with robotic control has led to the emergence of Vision-Language-Action (VLA) models, which have significantly advanced the field of embodied AI. Early pioneering works, such as RT-1~\citep{brohan2022rt} and RT-2~\citep{zitkovich2023rt}, demonstrated that pre-trained Vision-Language Models (VLMs) could be effectively repurposed to map raw visual observations and natural language instructions directly into low-level robot actions. Following this paradigm, open-source models like OpenVLA~\citep{kim2024openvla} and Octo~\citep{team2024octo} have become standard baselines, leveraging powerful 2D vision backbones (e.g., SigLIP~\citep{zhai2023sigmoid}, DINOv2~\citep{oquab2023dinov2}) and large-scale cross-embodiment datasets~\citep{o2024open} to achieve remarkable zero-shot generalization. More recently, architectures such as $\pi_0$~\citep{black2024pi_0, black2025pi_} and GR00T~\citep{bjorck2025gr00t} have introduced flow-matching and continuous action generation, further improving dexterous manipulation capabilities.
Despite their semantic reasoning capabilities, the visual backbones of these mainstream VLAs are primarily pretrained on massive 2D image-text pairs via contrastive learning. Consequently, they excel at capturing 2D semantic correspondences but inherently lack 3D spatial awareness~\citep{huang2025mllms}. This absence of geometric priors severely limits their performance in fine-grained manipulation tasks that require precise depth estimation, obstacle avoidance, and spatial relationship reasoning.

\subsection{Spatial Grounding for VLA Models}
To address the lack of 3D spatial understanding, recent works explore explicit and implicit spatial grounding. Explicit methods augment VLA inputs with depth maps, point clouds, or depth estimations~\citep{qu2025spatialvla, li2026pointvla, sun2025geovla}. However, these approaches suffer from hardware dependency, susceptibility to sensor noise, prediction error propagation, and significant computational overhead during inference. To circumvent these issues, concurrent works pursue implicit spatial grounding by leveraging 3D-aware foundation models, typically VGGT~\citep{wang2025vggt}. For instance, Evo-0~\citep{lin2025evo} integrates VGGT features via cross-attention, while Spatial Forcing~\citep{li2025spatial}, ROCKET~\citep{sun2026rocket}, and GLaD~\citep{guo2025glad} employ representation alignment at intermediate vision encoder layers or within the LLM hidden states. While promising, these implicit methods share critical limitations. They require exhaustive empirical searches to identify optimal alignment layers, which vary across tasks. More importantly, aligning at deep intermediate layers or within the LLM lacks geometric interpretability, as visual representations at these stages are already heavily entangled with linguistic semantics. In contrast, VEGA addresses these limitations by aligning directly at the output of the visual encoder. Instead of relying on VGGT~\citep{wang2025vggt}, VEGA utilizes FiT3D~\citep{yue2024improving}, which distills 3D Gaussian Splatting properties into a DINOv2 backbone, yielding dense geometric representations that perfectly match standard ViT architectures. By grounding spatial awareness at the purely visual stage before any linguistic entanglement occurs, VEGA offers a simpler, more interpretable, and highly effective spatial alignment framework without requiring complex layer searches or introducing inference overhead.

\section{Method}
In this section, we present the VEGA framework in detail. Sec.~\ref{sec:preliminaries} introduces the preliminaries on VLA models and the FiT3D visual encoder fine-tuning framework. Sec.~\ref{sec:motivation} provides motivation through feature visualization and controlled pretraining experiments, demonstrating the benefit of DINOv2-FiT3D features for robotic manipulation. Sec.~\ref{sec:alignment} describes the proposed visual encoder spatial alignment framework.

\subsection{Preliminaries}\label{sec:preliminaries}
\paragraph{Vision-Language-Action Models.}
VLA models extend pretrained vision-language models (VLMs) to robotic control by jointly processing visual observations, natural language instructions, and action signals. Given a set of multi-view images $\mathcal{I} = \{I_k\}_{k=1}^{K}$ captured by robot-mounted cameras and a language instruction $\mathcal{L}$, the visual observations are first tokenized into $N$ visual tokens $\{x^V_i\}_{i=1}^{N}$ through a pretrained visual encoder, while the instruction is converted into $M$ linguistic tokens $\{x^L_j\}_{j=1}^{M}$ via a text tokenizer. The VLA model then autoregressively generates $T$ action tokens $\{x^A_t\}_{t=1}^{T}$ conditioned on the preceding visual and linguistic context:
\begin{equation}
    x^A_t \sim p_\theta\left(x^A_t \mid \{x^V_i\}_{i=1}^{N}, \{x^L_j\}_{j=1}^{M}, x^A_{<t}\right),
\end{equation}
where $\theta$ denotes the parameters of the VLA backbone.  It is trained via an action prediction loss:
\begin{equation}
    \mathcal{L}_{\text{action}} = \mathcal{L}\left[\mathcal{G}(\{x^A_t\}_{t=1}^{T}),\ A_{\text{gt}}\right],
\end{equation}
where $\mathcal{G}$ denotes the action expert head (e.g., a two-layer MLP or a flow-matching head) and $A_{\text{gt}}$ denotes the ground-truth action sequence from demonstrations. Since visual tokens serve as the primary scene representation for action prediction, the quality of spatial information encoded in these tokens directly affects manipulation performance.

\paragraph{Visual Encoder 3D-Aware Fine-Tuning.}
FiT3D~\citep{yue2024improving} is a two-stage framework that injects 3D spatial awareness into 2D visual foundation models such as DINOv2~\citep{oquab2023dinov2} by leveraging multi-view consistent 3D representations. In the first stage, for each training scene, a set of multi-view images $\{I_i\}_{i=1}^{N}$ with corresponding camera poses is used to train a 3D Gaussian representation $\mathcal{G}$~\citep{kerbl20233d} that jointly optimizes RGB reconstruction and feature distillation. Each Gaussian is equipped with a low-dimensional feature vector $\mathbf{f} \in \mathbb{R}^D$, which is rendered into a feature image $\mathbf{F}^{\text{low}}$ via differentiable $\alpha$-blending and subsequently up-projected to the full feature dimension through a lightweight CNN decoder. The Gaussian geometry is supervised by RGB photometric loss, while the feature vectors are supervised by the 2D foundation features extracted from DINOv2, ensuring multi-view consistency is enforced through the 3D representation. In the second stage, the rendered 3D-aware features $\mathbf{F}^{\text{high}}$ from the pre-trained Gaussian representations serve as supervision targets to fine-tune the 2D foundation model with an $\ell_1$ loss:
\begin{equation}
    \mathcal{L}_{\text{FiT3D}} = \left\| \varepsilon^{\text{2D}}(I_i) - \mathbf{F}^{\text{high}}_i \right\|_1,
\end{equation}
where $\varepsilon^{\text{2D}}$ denotes the 2D feature extractor being fine-tuned. By aggregating features from multiple viewpoints into a geometrically consistent 3D representation and distilling them back into the 2D model, FiT3D produces dense patch-level features that are simultaneously spatially consistent and semantically rich, making it a natural candidate for a spatial teacher in VLA alignment.

\begin{figure}[!t]
  \centering
  \includegraphics[width=\textwidth]{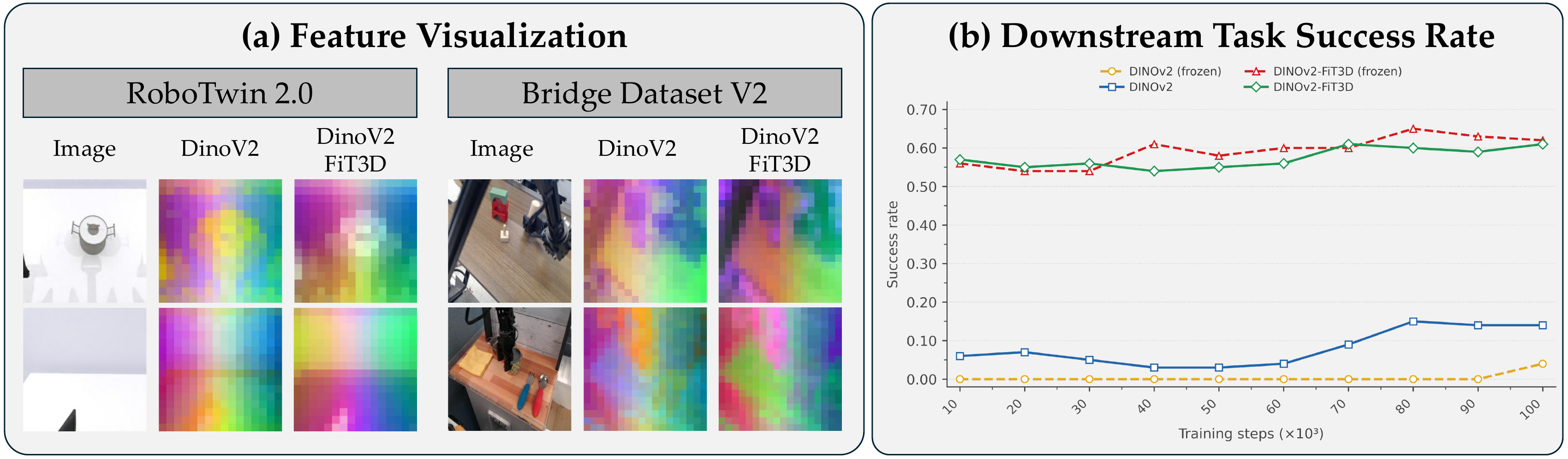}
  \caption{\textbf{DINOv2-FiT3D features exhibit stronger spatial structure and improve VLA manipulation performance.} \textbf{(a)} PCA visualizations comparing DINOv2 and DINOv2-FiT3D patch features, where FiT3D produces more spatially consistent representations with cleaner object boundaries. \textbf{(b)} Success rate on Move Playingcard Away for four encoder variants, showing that DINOv2-FiT3D substantially outperforms DINOv2 regardless of whether the encoder is frozen.}
  \label{fig:probe}
  \vspace{-1em}
\end{figure}

\subsection{Motivation}\label{sec:motivation}
\paragraph{DINOv2-FiT3D Features Transfer to Robotic Domains.}
Although DINOv2-FiT3D is fine-tuned exclusively on indoor scene datasets~\citep{yeshwanth2023scannet++} without any robotic data, its 3D-aware features retain strong spatial transferability to robotic manipulation scenarios. We conduct a feature visualization experiment on RoboTwin 2.0~\citep{chen2025robotwin} and Bridge Dataset v2~\citep{walke2023bridgedata}, comparing PCA visualizations of the original DINOv2 and DINOv2-FiT3D. As shown in Fig.~\ref{fig:probe}, DINOv2-FiT3D produces more compact and spatially consistent feature maps, with cleaner object boundaries and more structured representations of manipulation-relevant regions. This demonstrates that the 3D spatial awareness injected by FiT3D transfers effectively to the robotic domain, motivating its use as a spatial supervision signal for VLA training.

\paragraph{DINOv2-FiT3D Encoder Improves VLA Spatial Grounding.}
To further validate the benefit of DINOv2-FiT3D features, we conduct a controlled pretraining experiment within the OpenVLA framework~\citep{kim2024openvla}, pretraining four model variants on Bridge Dataset v2~\citep{walke2023bridgedata} and evaluating on Move Playingcard Away after fine-tuning. As shown in Fig.~\ref{fig:probe}, both DINOv2-FiT3D variants substantially outperform their DINOv2 counterparts, confirming that richer spatial features at the encoder level directly benefit downstream manipulation. The negligible gap between frozen and unfrozen DINOv2-FiT3D variants further indicates high intrinsic feature quality without task-specific adaptation. However, directly replacing the visual encoder and retraining from scratch incurs prohibitive computational cost, motivating our alignment-based approach, VEGA, which efficiently transfers the spatial awareness of DINOv2-FiT3D into an existing VLA model.

\begin{figure}[!t]
  \centering
  \includegraphics[width=\textwidth]{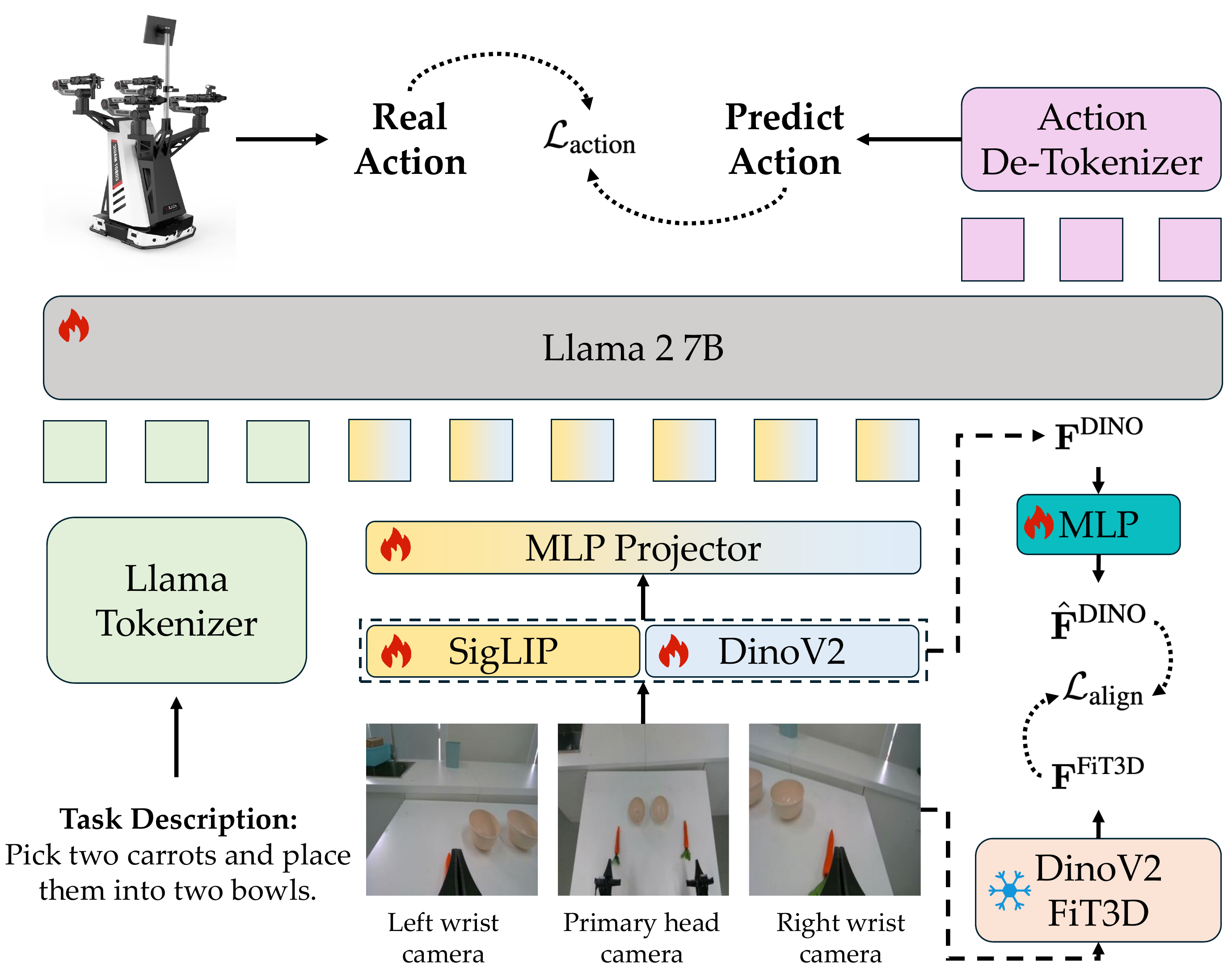}
  \caption{\textbf{Overview of the VEGA training framework.} The frozen DINOv2-FiT3D encoder serves as a spatial teacher, supervising the DINOv2 branch of the VLA visual backbone via a lightweight projector. The alignment loss is computed as the cosine distance between projected DINOv2 features and FiT3D features, and is combined with the action prediction loss during training. At inference time, the projector and teacher encoder are discarded, introducing no additional overhead.}
  \label{fig:method}
  \vspace{-1em}
\end{figure}

\subsection{Visual Encoder Spatial Alignment}\label{sec:alignment}
Given a pretrained VLA model whose visual backbone consists of a DINOv2 encoder $\varepsilon^{\text{DINO}}$ and a SigLIP encoder $\varepsilon^{\text{SigLIP}}$, VEGA performs spatial alignment exclusively on the DINOv2 branch, as it is responsible for fine-grained spatial and structural representations. A frozen DINOv2-FiT3D encoder $\varepsilon^{\text{FiT3D}}$ serves as the spatial teacher. Formally, given an input image $I$, we extract patch-level visual tokens from the second-to-last transformer block of $\varepsilon^{\text{DINO}}$, bypassing FiLM language conditioning~\citep{perez2018film} to preserve vision-specific representations, consistent with OpenVLA's standard visual backbone design~\citep{kim2024openvla, kim2025fine}. For the teacher, we extract features from the final transformer block of $\varepsilon^{\text{FiT3D}}$, which contains the most complete 3D spatial representations. This asymmetric design follows a common practice in knowledge distillation: using the teacher's strongest signal to supervise the student's working layer~\citep{chen2022knowledge}:
\begin{equation}
\mathbf{F}^{\text{DINO}} = \varepsilon^{\text{DINO}}_{L-2}(I), \quad \mathbf{F}^{\text{FiT3D}} = \varepsilon^{\text{FiT3D}}_{L-1}(I),
\end{equation}
where $\mathbf{F}^{\text{DINO}}, \mathbf{F}^{\text{FiT3D}} \in \mathbb{R}^{N \times d}$ denote the patch token features with $N$ tokens and feature dimension $d=1024$. To perform a non-linear transformation into a better-aligned feature space, we introduce a lightweight projector $\phi$ consisting of a LayerNorm and a two-layer MLP with GELU activation:
\begin{equation}
\hat{\mathbf{F}}^{\text{DINO}} = \phi(\mathbf{F}^{\text{DINO}}).
\end{equation}
The alignment loss is computed as the average cosine distance between the projected DINOv2 features and the DINOv2-FiT3D features:
\begin{equation}
\mathcal{L}_{\text{align}} = \frac{1}{N}\sum_{i=1}^{N} \left(1 - \frac{\hat{\mathbf{F}}^{\text{DINO}}_i \cdot \mathbf{F}^{\text{FiT3D}}_i}{\|\hat{\mathbf{F}}^{\text{DINO}}_i\| \|\mathbf{F}^{\text{FiT3D}}_i\|}\right).
\end{equation}
The final training objective combines the action prediction loss with the alignment loss weighted by a scalar $\lambda$:
\begin{equation}
\mathcal{L}_{\text{VEGA}} = \mathcal{L}_{\text{action}} + \lambda \mathcal{L}_{\text{align}},
\end{equation}
where $\lambda$ balances the contribution of spatial alignment against task learning. During inference, VEGA introduces no additional computational overhead, as the projector and $\varepsilon^{\text{FiT3D}}$ are discarded, and the VLA model operates identically to its standard form.

\section{Experiment}\label{sec:experiments}

In this section, we comprehensively evaluate VEGA across simulation and real-world settings. Sec.~\ref{sec:setup} describes the experimental setup. Sec.~\ref{sec:sim} presents quantitative comparisons against baselines on RoboTwin 2.0 and analyzes training and data efficiency. Sec.~\ref{sec:ablation} conducts ablation studies on key design choices, including the alignment loss coefficient and the spatial teacher model. Sec.~\ref{sec:real} validates VEGA on a physical bimanual ALOHA platform across tasks of varying complexity.

\subsection{Experimental Setup}\label{sec:setup}
\paragraph{Simulation Environment.}
We evaluate VEGA on the RoboTwin 2.0 benchmark~\citep{chen2025robotwin} using the AgiLeX ALOHA platform, covering six bimanual manipulation tasks: \textit{Move Playingcard Away}, \textit{Turn Switch}, \textit{Click Bell}, \textit{Beat Block}, \textit{Lift Pot}, and \textit{Place Shoes}, following the evaluation protocol of~\citep{li2025spatial, sun2026rocket}. We train on the Easy setting using the official clean demonstration data and evaluate on both Easy and Hard settings, where the Hard setting introduces domain randomization, including scene clutter, diverse background textures, lighting variation, and varied tabletop heights. Each task is evaluated over 100 trials. We report task success rate as the evaluation metric.

\paragraph{Real-World Settings.}
We validate VEGA on a physical bimanual AgileX ALOHA platform across four tasks of varying complexity: two single-arm tasks (\textit{Close Laptop} and \textit{Handover Cucumber}) and two bimanual tasks (\textit{Pick Dual Carrots into Dual Bowls} and \textit{Pick Dual Flowers into Vase} ). For each task, we collect 100 demonstration trajectories via teleoperation and fine-tune the model following the same training protocol as in simulation. Each task is evaluated over 20 trials, and we report the success rate as the evaluation metric.

\paragraph{Implementation Details.}
We build VEGA upon OpenVLA-OFT~\citep{kim2025fine} as the base VLA model. The alignment projector consists of a two-layer MLP with GELU activation, projecting the VLA visual encoder output tokens into the DINOv2-FiT3D feature space of dimension 1024, with a LayerNorm applied to the input features. The DINOv2-FiT3D encoder is kept frozen throughout training. We apply LoRA~\citep{hu2022lora} fine-tuning with rank 32 to the LLM backbone. All models are trained for 100k steps with a batch size of 4 per GPU on 4 NVIDIA H100 GPUs, with a learning rate of $5 \times 10^{-4}$ decayed after 50k steps. Each run takes approximately 28 hours. The alignment loss coefficient is set to $\lambda = 0.1$. We enable image augmentation, proprioception input, and FiLM~\citep{perez2018film} language conditioning following the OpenVLA-OFT training protocol. All baselines are reproduced under the same training configuration to ensure a fair comparison.

\subsection{Simlulation Experiments}\label{sec:sim}
\paragraph{Main Results.}
\begin{table*}[t]
\caption{\textbf{Quantitative comparison.} Performance across all manipulation tasks. \colorbox[HTML]{E2EFDA}{Green} indicates the best and \colorbox[HTML]{FFF2CC}{yellow} indicates the second-best performance. AVG is computed over valid entries.}
\centering
\footnotesize
\renewcommand{\arraystretch}{1.2}
\setlength{\tabcolsep}{3pt}
\begin{tabular}{l|cc|cc|cc|cc|cc|cc|cc}
\toprule
\textbf{Method}
  & \multicolumn{2}{c|}{\shortstack{Move Card\\Away}}
  & \multicolumn{2}{c|}{\shortstack{Turn\\Switch}}
  & \multicolumn{2}{c|}{\shortstack{Click\\Bell}}
  & \multicolumn{2}{c|}{\shortstack{Beat\\Block}}
  & \multicolumn{2}{c|}{\shortstack{Lift\\Pot}}
  & \multicolumn{2}{c|}{\shortstack{Place\\Shoe}}
  & \multicolumn{2}{c}{\textbf{AVG}} \\
\cmidrule(lr){2-3}\cmidrule(lr){4-5}\cmidrule(lr){6-7}
\cmidrule(lr){8-9}\cmidrule(lr){10-11}\cmidrule(lr){12-13}\cmidrule(lr){14-15}
& Easy & Hard & Easy & Hard & Easy & Hard
& Easy & Hard & Easy & Hard & Easy & Hard
& Easy & Hard \\
\midrule
DP~\citep{chi2025diffusion}
  & 0.47 & 0.00
  & 0.36 & 0.01
  & 0.54 & 0.00
  & 0.42 & 0.19
  & 0.39 & 0.08
  & 0.08 & 0.01
  & 0.377 & 0.048 \\
RDT~\citep{liu2024rdt}
  & 0.43 & 0.11
  & 0.35 & 0.15
  & 0.79 & 0.08
  & \cellcolor[HTML]{E2EFDA}0.77 & \cellcolor[HTML]{E2EFDA}0.37
  & 0.71 & 0.09
  & 0.04 & 0.04
  & 0.515 & 0.140 \\
$\pi_0$~\citep{black2024pi_0}
  & 0.53 & 0.22
  & 0.27 & \cellcolor[HTML]{FFF2CC}0.23
  & 0.42 & 0.02
  & 0.43 & 0.21
  & 0.82 & 0.38
  & 0.15 & 0.00
  & 0.437 & 0.177 \\
$\pi_0$ + SF~\citep{li2025spatial}
  & 0.60 & 0.33
  & \cellcolor[HTML]{FFF2CC}0.38 & \cellcolor[HTML]{E2EFDA}0.24
  & 0.46 & 0.07
  & 0.46 & \cellcolor[HTML]{FFF2CC}0.35
  & 0.90 & \cellcolor[HTML]{FFF2CC}0.41
  & 0.21 & 0.03
  & 0.502 & 0.238 \\
$\pi_0$ + ROCKET~\citep{sun2026rocket}
  & 0.73 & \cellcolor[HTML]{FFF2CC}0.35
  & 0.25 & 0.21
  & {---} & {---}
  & \cellcolor[HTML]{FFF2CC}0.50 & 0.04
  & {---} & {---}
  & 0.25 & 0.00
  & 0.433 & 0.150 \\
OpenVLA-OFT~\citep{kim2025fine}
  & 0.70 & 0.34
  & 0.20 & 0.08
  & \cellcolor[HTML]{FFF2CC}0.95 & 0.46
  & 0.02 & 0.06
  & \cellcolor[HTML]{E2EFDA}0.98 & 0.33
  & \cellcolor[HTML]{FFF2CC}0.51 & 0.09
  & 0.560 & 0.227 \\
OFT + SF~\citep{li2025spatial}
  & \cellcolor[HTML]{FFF2CC}0.76 & 0.32
  & \cellcolor[HTML]{E2EFDA}0.39 & 0.20
  & \cellcolor[HTML]{E2EFDA}0.97 & \cellcolor[HTML]{FFF2CC}0.48
  & 0.36 & 0.10
  & \cellcolor[HTML]{FFF2CC}0.91 & \cellcolor[HTML]{E2EFDA}0.44
  & 0.46 & \cellcolor[HTML]{FFF2CC}0.13
  & \cellcolor[HTML]{FFF2CC}0.642 & \cellcolor[HTML]{FFF2CC}0.278 \\
\midrule
Ours
  & \cellcolor[HTML]{E2EFDA}0.77 & \cellcolor[HTML]{E2EFDA}0.43
  & \cellcolor[HTML]{E2EFDA}0.39 & \cellcolor[HTML]{E2EFDA}0.24
  & \cellcolor[HTML]{E2EFDA}0.97 & \cellcolor[HTML]{E2EFDA}0.53
  & 0.38 & 0.11
  & \cellcolor[HTML]{E2EFDA}0.98 & 0.28
  & \cellcolor[HTML]{E2EFDA}0.56 & \cellcolor[HTML]{E2EFDA}0.25
  & \cellcolor[HTML]{E2EFDA}0.675 & \cellcolor[HTML]{E2EFDA}0.307 \\
\bottomrule
\end{tabular}
\label{tab:final_results}
\end{table*}
Tab.~\ref{tab:final_results} presents the quantitative comparison on RoboTwin 2.0 across six bimanual manipulation tasks. VEGA achieves the highest average success rate of 67.5\% on Easy and 30.7\% on Hard, outperforming the strongest baseline OFT+SF by 3.3\% and 2.9\%, respectively. Compared to the base model OpenVLA-OFT, VEGA brings consistent improvements across most tasks, with particularly notable gains on Move Card Away, Turn Switch, and Place Shoes, suggesting that tasks requiring precise spatial localization benefit most from the injected 3D spatial awareness. On the Hard setting, where domain randomization is more severe, VEGA shows the most pronounced advantage over all baselines, indicating that visual encoder-level spatial grounding provides more robust representations under distribution shift.

\begin{figure}[!t]
  \centering
  \includegraphics[width=\textwidth]{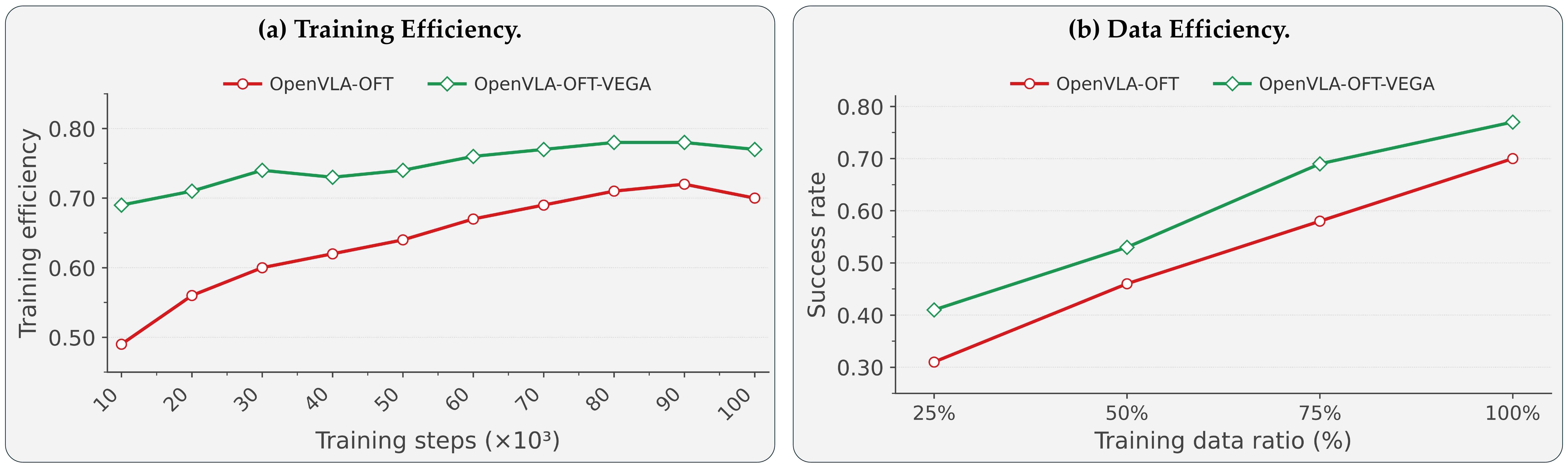}
  \caption{\textbf{VEGA improves both training and data efficiency over OpenVLA-OFT.} \textbf{(a)} Success rate curves across training steps on Move Playingcard Away (Easy), showing that VEGA converges faster and reaches a higher performance ceiling. \textbf{(b)} Success rate under varying demonstration data fractions, demonstrating that VEGA consistently outperforms the base model across all data regimes.}
  \label{fig:efficiency}
\end{figure}

\paragraph{Training Efficiency.}
To assess the training efficiency of VEGA, we compare the learning curves of OpenVLA-OFT and VEGA on the Move Playingcard Away task under the Easy setting. As shown in Fig.~\ref{fig:efficiency}(a), VEGA consistently outperforms the base model throughout training. Notably, VEGA achieves at 10k steps a success rate comparable to that of OpenVLA-OFT at 60k steps, demonstrating that visual encoder-level spatial alignment significantly accelerates policy learning. Both models converge around 80k--90k steps, but VEGA converges to a higher plateau, suggesting that the injected spatial awareness not only speeds up learning but also improves the final performance ceiling.

\paragraph{Data Efficiency.}
To evaluate the data efficiency of VEGA, we train both OpenVLA-OFT and VEGA on varying fractions of the demonstration data (25\%, 50\%, 75\%, and 100\%) on the Move Playingcard Away task and report success rates under the Easy setting. As shown in Fig.~\ref{fig:efficiency}(b), VEGA consistently outperforms OpenVLA-OFT across all data regimes. Under the most data-scarce setting, VEGA achieves a 10\% absolute improvement over the base model, demonstrating that spatial alignment provides a strong inductive bias that partially compensates for limited demonstration data. This performance gap persists across all data fractions, indicating that VEGA makes more effective use of available training data at every scale.

\subsection{Ablation Study}\label{sec:ablation}

\paragraph{Effect of Alignment Loss Coefficient $\lambda$.}
\begin{figure}[!t]
  \centering
  \includegraphics[width=\textwidth]{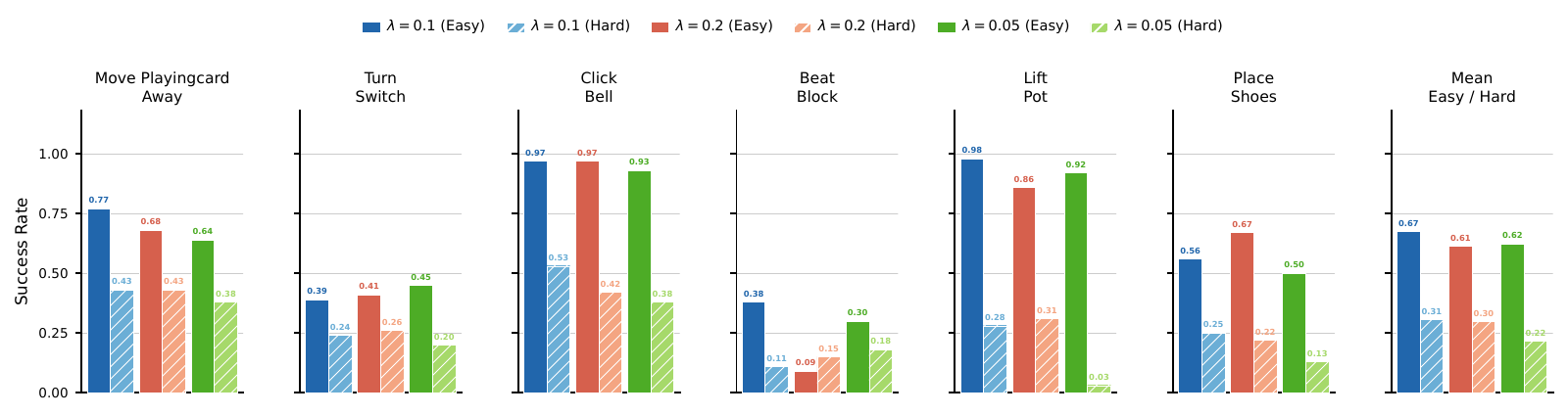}
    \caption{\textbf{Effect of alignment loss coefficient $\lambda$.} Success rate (\%) on RoboTwin 2.0 across six tasks under Easy and Hard settings with $\lambda \in \{0.05, 0.1, 0.2\}$, where $\lambda=0.1$ achieves the best overall balance between spatial alignment and task learning.}
\label{fig:ablation}
\vspace{-1em}
\end{figure}

We ablate the alignment loss coefficient $\lambda$, which controls the balance between action prediction and spatial alignment supervision. As shown in Fig.~\ref{fig:ablation}, $\lambda=0.1$ achieves the best overall performance, with an average success rate of 67.5\% on Easy and 30.7\% on Hard. Increasing $\lambda$ to 0.2 leads to a notable drop on Easy (61.3\%), suggesting that excessive alignment supervision interferes with task learning. Decreasing $\lambda$ to 0.05 similarly degrades performance, particularly on Hard (21.7\%), indicating insufficient spatial grounding transfer. These results confirm that $\lambda=0.1$ strikes the best balance between spatial alignment and action prediction, and we adopt it as the default setting in all experiments.

\begin{wraptable}{r}{0.35\textwidth}
\caption{\textbf{Ablation on spatial teacher model.} Success rate on Move Playingcard Away under Easy and Hard settings.}
\centering
\footnotesize
\renewcommand{\arraystretch}{1.2}
\setlength{\tabcolsep}{4pt}
\begin{tabular}{l|cc}
\toprule
\textbf{Method} & \textbf{Easy} & \textbf{Hard} \\
\midrule
w/o teacher        & 0.70 & 0.34 \\
VGGT as teacher    & 0.76 & 0.04 \\
FiT3D as teacher (ours) & \textbf{0.77} & \textbf{0.43} \\
\bottomrule
\end{tabular}
\label{tab:ablation_teacher}
\vspace{-1em}
\end{wraptable}
\paragraph{Effect of Spatial Teacher Model.}
Tab.~\ref{tab:ablation_teacher} compares three variants for the choice of spatial teacher. Removing the teacher entirely yields the lowest Easy performance, confirming the benefit of spatial alignment. Replacing the teacher with VGGT-at-encoder yields a marginal Easy improvement but catastrophically degrades Hard performance (0.04), falling far below even the no-teacher baseline. We attribute this to an architectural mismatch: VGGT's intermediate representations encode scene-level geometry optimized for novel view synthesis rather than dense patch-level features compatible with ViT token alignment, and this mismatch destabilizes generalization under domain shift. In contrast, VEGA with DINOv2-FiT3D achieves the best performance on both settings, particularly on Hard (0.43 vs.\ 0.04), owing to FiT3D's architecturally homogeneous patch-level features derived from multi-view consistent 3D Gaussian Splatting supervision.

\subsection{Real-World Experiments}\label{sec:real}
\begin{figure}[!t]
  \centering
  \includegraphics[width=\textwidth]{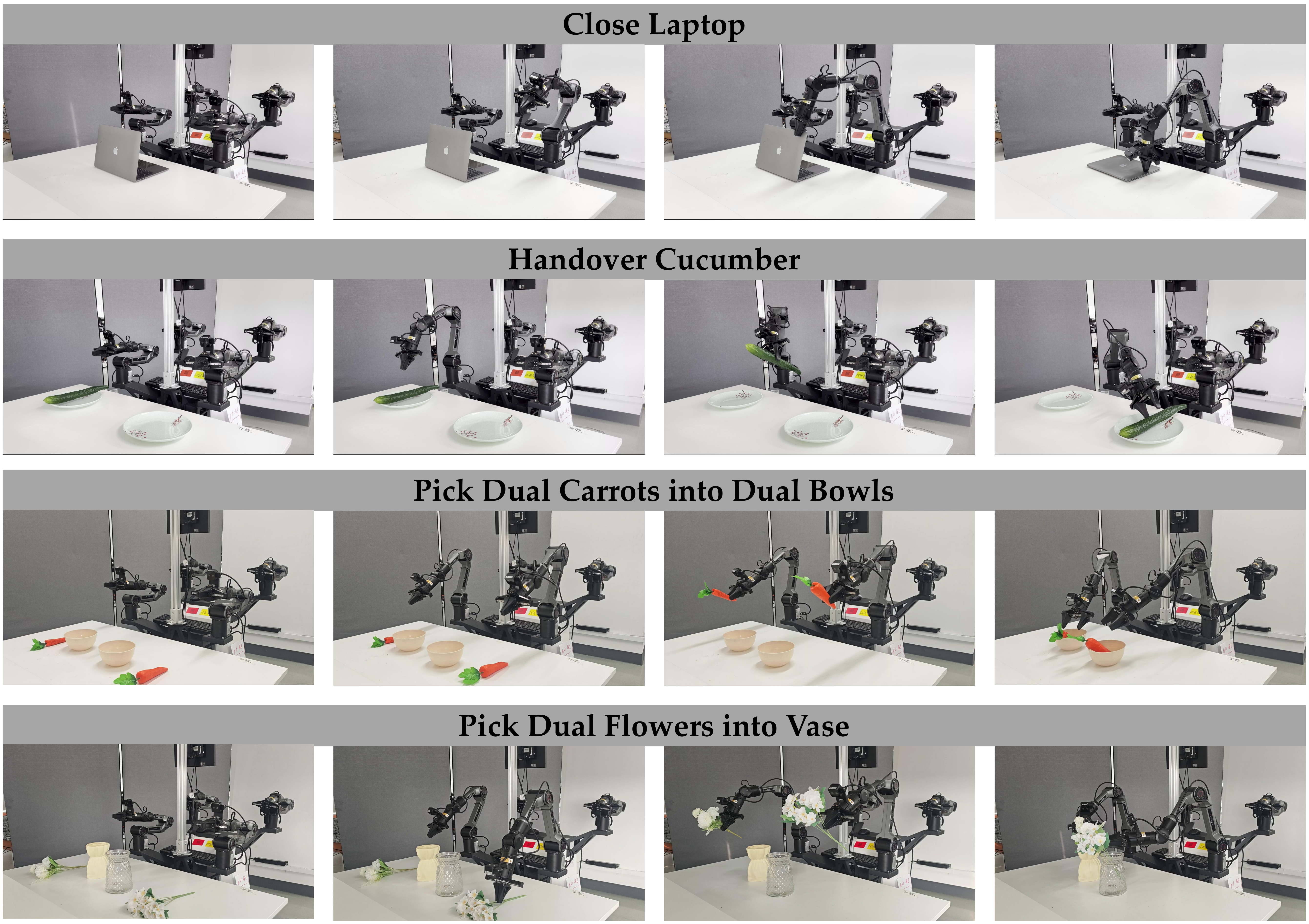}
  \caption{\textbf{Qualitative visualization of VEGA on real-world manipulation tasks.} Keyframe sequences illustrating the execution of four tasks: \textit{Close Laptop}, \textit{Handover Cucumber}, \textit{Pick Dual Carrots into Dual Bowls}, and \textit{Pick Dual Flowers into Vase}. Each row shows the progression from the initial state to task completion.}
  \label{fig:realdemo}
\end{figure}
\begin{table*}[t]
\caption{\textbf{Quantitative comparison of real-world experiments.} Success rates across single-arm and bimanual manipulation tasks. }
\centering
\footnotesize
\renewcommand{\arraystretch}{1.2}

\newcolumntype{C}{>{\centering\arraybackslash}p{2cm}}

\begin{tabular}{l | C C | C C | c}
\toprule
\multirow{2}{*}{\textbf{Method}}
  & \multicolumn{2}{c|}{\textbf{Single-Arm Tasks}}
  & \multicolumn{2}{c|}{\textbf{Bimanual Tasks}}
  & \multirow{2}{*}{\textbf{AVG}} \\
\cmidrule(lr){2-3}\cmidrule(lr){4-5}
  & \shortstack{Close\\Laptop} 
  & \shortstack{Handover\\Cucumber} 
  & \shortstack{Pick Dual\\Carrots} 
  & \shortstack{Pick Dual\\Flowers} 
  & \\
\midrule
OpenVLA-OFT~\citep{kim2025fine}
  & 0.65 & 0.60 & 0.50 & 0.15 & 0.48 \\
OFT + SF~\citep{li2025spatial}
  & 0.70 & 0.70 & \textbf{0.60} & 0.20 & 0.55 \\
\midrule
Ours
  & \textbf{0.80} & \textbf{0.75} & \textbf{0.60} & \textbf{0.25} & \textbf{0.60} \\
\bottomrule
\end{tabular}
\label{tab:real_world_results}
\end{table*}

\paragraph{Result Analysis.}
Tab.~\ref{tab:real_world_results} presents the quantitative comparison on real-world manipulation tasks, with qualitative execution visualizations shown in Fig.~\ref{fig:realdemo}. VEGA consistently outperforms both OpenVLA-OFT and OFT+SF across all four tasks, demonstrating that visual encoder-level spatial grounding transfers effectively to physical robot settings. The performance gains are most pronounced on spatially demanding tasks such as \textit{Close Laptop} and \textit{Pick Dual Flowers into Vase}. Since these tasks require reasoning about 3D articulated structures and precise depth estimation for narrow insertions, the fine-grained geometric awareness preserved by VEGA at the visual encoder stage proves critical. On \textit{Pick Dual Carrots into Dual Bowls}, VEGA matches OFT+SF while outperforming the base model, likely because the larger target receptacles (bowls) offer higher spatial tolerance, reducing the need for millimeter-level precision. Overall, these results validate that VEGA's spatial grounding generalizes robustly to real-world manipulation.

\section{Conclusion}
In this work, we presented VEGA, a simple yet effective framework for implicit spatial grounding in VLA models. By directly aligning visual encoder outputs with spatially-aware DINOv2-FiT3D features via a lightweight projector, VEGA grounds spatial awareness before any linguistic entanglement occurs, eliminating the need for empirical layer search while offering a more geometrically interpretable alignment target. The projector is discarded at inference time, introducing no additional computational overhead. Extensive experiments on RoboTwin 2.0 and real-world manipulation tasks demonstrate that VEGA consistently outperforms existing implicit spatial grounding baselines, achieving state-of-the-art performance while also improving training and data efficiency. One limitation is that VEGA currently relies on DINOv2-FiT3D as the spatial teacher, whose transferability may degrade in highly unstructured or out-of-distribution environments. Looking forward, exploring stronger 3D-aware teacher models and extending the visual encoder alignment paradigm to broader VLA architectures remain promising directions for future work.

\clearpage
\bibliography{references}{}
\bibliographystyle{plain}


\newpage
\appendix

\section{Additional Experimental Setup}
\subsection{Pretraining Details}
For the controlled pretraining experiment in Sec.~\ref{sec:motivation}, we initialize the model from the Prismatic-DINO-SigLIP-224px+7B base VLM and pretrain on the Bridge mixture from Open X-Embodiment~\citep{o2024open} using 4 GPUs with FSDP full-shard training. The per-device batch size is 32, resulting in a global batch size of 128. We use a constant learning rate of $2\times10^{-5}$ with no warmup, weight decay of 0.0, and gradient clipping with maximum norm 1.0. Gradient checkpointing and mixed-precision training are enabled, and image augmentation is disabled. Training is configured for up to 1000 epochs, but stopped early after 8 epochs upon convergence. For the FiT3D variants, the only modification is that the DINOv2 branch is initialized from a FiT3D checkpoint prior to pretraining.

\subsection{Baseline Method Details}

In this section, we provide detailed descriptions of the baseline methods evaluated in our experiments.

\subsubsection{Diffusion-based Policies}
\begin{itemize}
    \item \textbf{DP~\citep{chi2025diffusion}:} Diffusion Policy treats action generation as a conditional denoising process, learning to map observation sequences to multi-modal action distributions. It serves as our foundational baseline for continuous imitation learning.
    \item \textbf{RDT~\citep{liu2024rdt}:} Robotics Diffusion Transformer leverages a Transformer-based architecture to model robot trajectories, effectively scaling to multi-modal inputs and diverse manipulation tasks.
\end{itemize}

\subsubsection{Foundation Model Variants}
\begin{itemize}
    \item \textbf{$\pi_0$~\citep{black2024pi_0}:} A versatile vision-language-action (VLA) foundation model trained on large-scale robotic data, designed to unify language reasoning and continuous motor control.
    
    \item \textbf{$\pi_0$ + SF~\citep{li2025spatial}:} This baseline integrates Spatial Forcing (SF) with $\pi_0$. SF is a representation alignment strategy that implicitly forces the VLA model to develop spatial comprehension without relying on explicit 3D sensor inputs (e.g., depth maps) or depth estimators. It achieves this by aligning the intermediate visual embeddings of $\pi_0$ with geometric representations produced by pretrained 3D foundation models, thereby enhancing action precision in the 3D physical world.
    
    \item \textbf{$\pi_0$ + ROCKET~\citep{sun2026rocket}:} ROCKET improves upon standard representation alignment by introducing a residual-oriented multi-layer alignment framework. It utilizes a shared projector to align multiple layers of the VLA backbone with a powerful 3D vision foundation model, which efficiently captures rich spatial information distributed across depths while mitigating gradient interference. 
    
    \item \textbf{OpenVLA-OFT~\citep{kim2025fine}:} This approach adapts the open-source OpenVLA foundation model to specific manipulation tasks using Orthogonal Fine-Tuning (OFT). OFT provides a parameter-efficient fine-tuning method that preserves the model's pre-trained generalization capabilities while learning new task distributions.
    
    \item \textbf{OFT + SF~\citep{li2025spatial}:} This baseline applies the Spatial Forcing (SF) alignment strategy on top of the OpenVLA-OFT model. It evaluates whether the implicit 3D spatial alignment provided by SF can synergize with parameter-efficient fine-tuning to further improve manipulation accuracy.
\end{itemize}

\subsection{Projector Architecture Details}

The alignment projector $\phi$ is a lightweight two-layer MLP that maps VLA visual encoder output tokens into the DINOv2-FiT3D feature space. Specifically, given input features of dimension $d_{\text{in}} = 1024$, the projector applies an optional LayerNorm, followed by a linear layer projecting to the teacher feature dimension $d_{\text{out}} = 1024$, a GELU activation, and a second linear layer maintaining the same dimension:
\begin{equation}
    \phi(\mathbf{F}) = \mathbf{W}_2 \cdot \text{GELU}(\mathbf{W}_1 \cdot \text{LayerNorm}(\mathbf{F}) + \mathbf{b}_1) + \mathbf{b}_2,
\end{equation}
where $\mathbf{W}_1, \mathbf{W}_2 \in \mathbb{R}^{1024 \times 1024}$ and $\mathbf{b}_1, \mathbf{b}_2 \in \mathbb{R}^{1024}$. All linear layers are initialized with Xavier uniform initialization and zero biases. The projector contains approximately 2.1M parameters, introducing negligible overhead during training, and is fully discarded at inference time.

\subsection{Real-World Robot Hardware Setup}
In our real-world experiments, we employ the AgileX Cobot Magic platform equipped with four 6-DoF AgileX Piper robotic arms. Both the leader and follower arms are equipped with parallel grippers featuring an 85 mm stroke. The robotic arms operate under joint position control. The robot is integrated with three RGB-D cameras: an Intel RealSense D435 mounted on the head providing a global perspective for environmental perception, and two Intel RealSense D405 cameras attached to the left and right wrists, respectively, offering localized visual feedback for fine-grained manipulation tasks. All algorithms utilize RGB information from all three cameras.

\subsection{Details of Real-World Data Collection}
Building upon our robot hardware setup, we collect four real-world tasks, comprising two single-arm tasks and two bimanual tasks. For each task, 100 demonstrations were collected via leader--follower teleoperation. Each demonstration comprises time-synchronized recordings of the follower arm's joint positions and RGB video streams from three camera perspectives. To ensure data diversity, objects were placed in varying positions on the table. Detailed descriptions of the four tasks are as follows:

\begin{enumerate}
    \item \textbf{Close Laptop.} The robot uses a single arm to close an open laptop screen. This task requires precise spatial perception to locate the screen and hinge, as well as smooth and controlled motion to avoid damaging the articulated structure during contact.

    \item \textbf{Handover Cucumber.} The robot grasps a cucumber from one plate and places it onto another using a single arm. This task requires accurate object localization and a smooth transfer trajectory to ensure stable grasping and precise placement without dropping the object.

    \item \textbf{Pick Dual Carrots into Dual Bowls.} Each arm simultaneously grasps a carrot and places it into the nearest corresponding bowl. This bimanual task requires synchronized motion planning and spatial reasoning to correctly associate each carrot with its target bowl, while executing two independent manipulation sequences in parallel without inter-arm interference.
    
    \item \textbf{Pick Dual Flowers into Vase.} Each arm independently grasps a flower and inserts it into its corresponding vase. This bimanual task requires coordinated motion planning and spatial reasoning, as the robot must simultaneously manage two independent manipulation sequences while avoiding inter-arm interference.
\end{enumerate}

\section{Additional Analysis}
\subsection{Feature Representation Analysis}

\begin{figure}[!ht]
  \centering
  \includegraphics[width=0.9\textwidth]{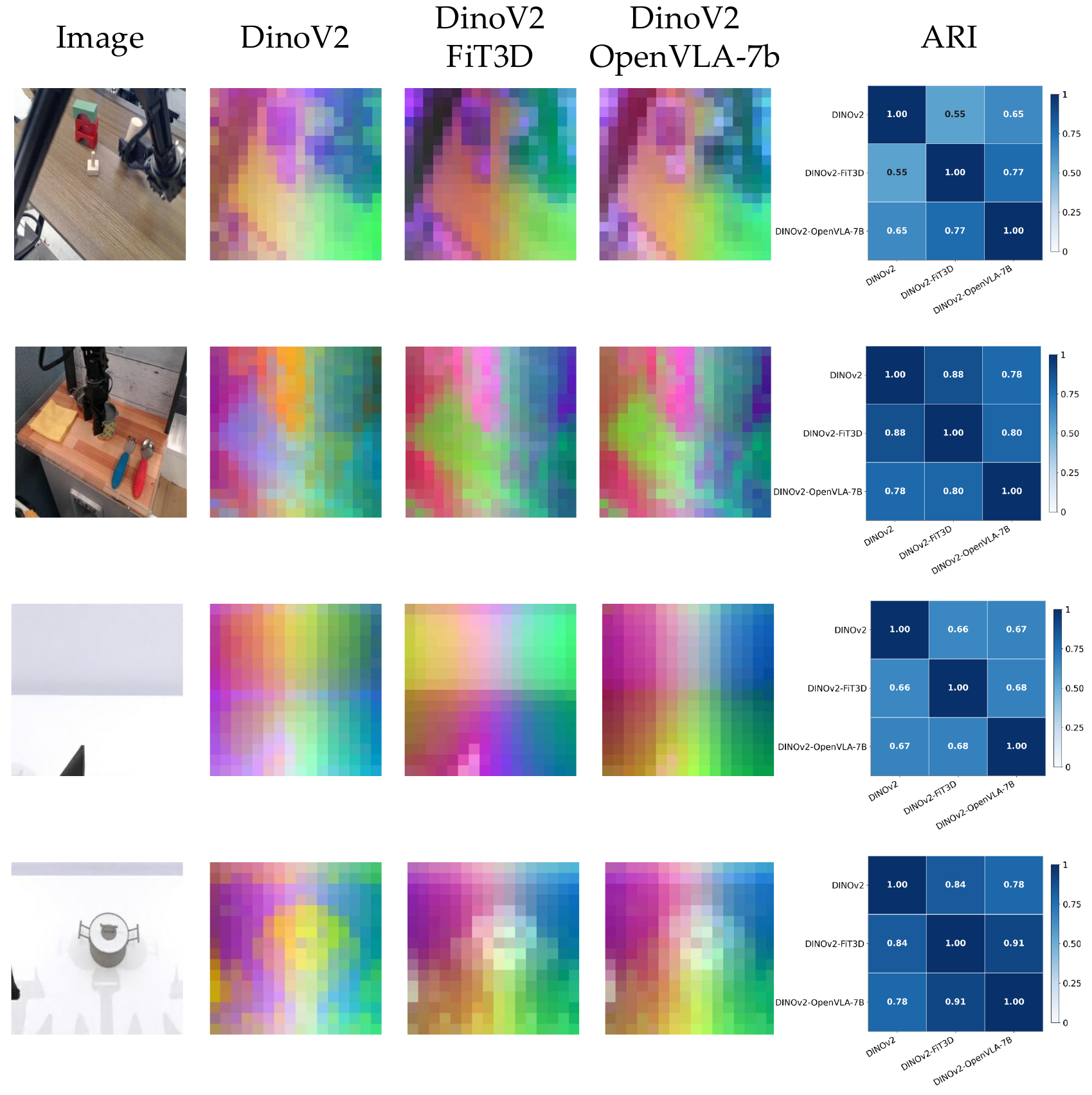}
  \caption{\textbf{Feature representation analysis across encoder variants.} Each row shows a scene from the robotic manipulation domain. Columns 2--4 show PCA visualizations of patch-level features from DINOv2, DINOv2-FiT3D, and DINOv2-OpenVLA-7B, respectively. Column 5 shows the pairwise ARI matrix computed from KMeans clustering of patch features, measuring the consistency of spatial groupings across encoders.}
  \label{fig:feat_vis}
\end{figure}

Fig.~\ref{fig:feat_vis} presents a qualitative and quantitative comparison of patch-level representations across three encoder variants: the original DINOv2, DINOv2-FiT3D, and the DINOv2 branch of OpenVLA-7B after VLA training.

The PCA visualizations reveal that DINOv2-FiT3D produces more spatially structured feature maps with cleaner object boundaries compared to the original DINOv2, particularly in scenes with complex foreground objects. Notably, the DINOv2-OpenVLA-7B features exhibit a visual structure closer to DINOv2-FiT3D than to the original DINOv2, suggesting that VLA training itself partially shifts the encoder toward more spatially organized representations.

The ARI matrices quantitatively corroborate this observation: DINOv2-FiT3D and DINOv2-OpenVLA-7B consistently achieve higher mutual ARI than either does with the original DINOv2, indicating that their spatial groupings are more consistent with each other. This suggests that the representational space toward which VLA training naturally evolves is geometrically closer to what FiT3D explicitly supervises, providing additional justification for using DINOv2-FiT3D as the spatial teacher in VEGA.

\subsection{Pretraining Convergence Analysis}

\begin{figure}[!t]
  \centering
  \includegraphics[width=\textwidth]{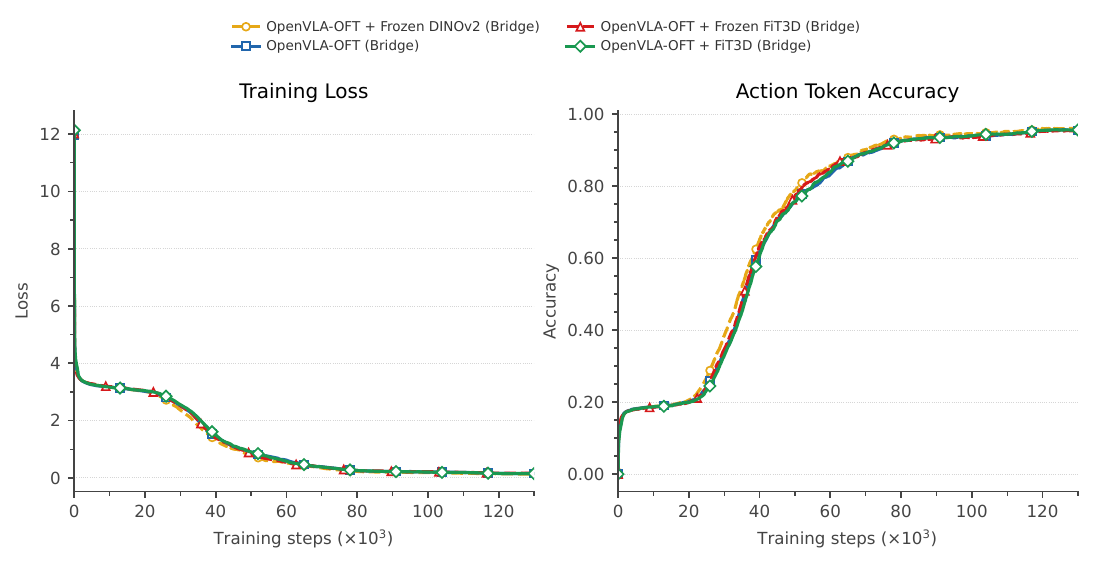}
  \caption{\textbf{Pretraining convergence curves.} Training loss and action token accuracy for OpenVLA-OFT and the FiT3D variant pretrained on Bridge Dataset v2. Both variants converge at a similar rate and to comparable final values, indicating that initializing the DINOv2 branch from a FiT3D checkpoint introduces no instability or degradation to the pretraining process.}
  \label{fig:pretrain_curve}
\end{figure}

Fig.~\ref{fig:pretrain_curve} shows the training loss and action token accuracy curves for OpenVLA-OFT and the FiT3D variant during Bridge pretraining. The two curves are nearly identical throughout training, converging at the same rate and reaching comparable final values. This confirms that replacing the DINOv2 encoder with a FiT3D checkpoint introduces no optimization instability or pretraining degradation, and that the performance gains observed after fine-tuning are attributable solely to the richer spatial representations of FiT3D rather than any difference in pretraining dynamics.

\begin{figure}[!t]
  \centering
  \includegraphics[width=\textwidth]{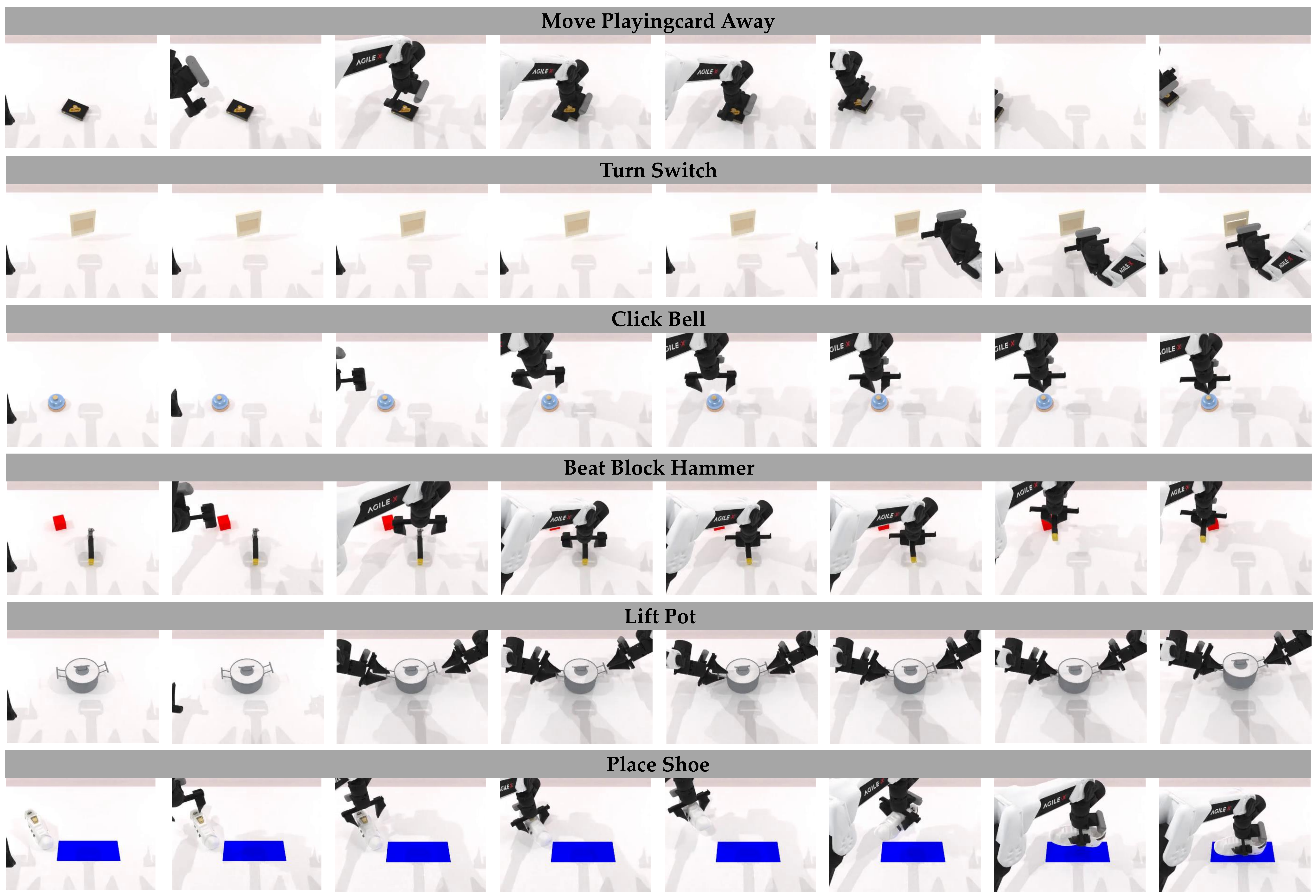}
  \caption{\textbf{Qualitative visualization of VEGA on RoboTwin 2.0.} Keyframe sequences illustrating the execution of all six bimanual manipulation tasks: Move Playingcard Away, Turn Switch, Click Bell, Beat Block, Lift Pot, and Place Shoes. Each row shows the progression from the initial state to task completion.}
  \label{fig:sim_vis}
\end{figure}

\section{Additional Experiment Results}

\subsection{Impact of Visual Encoder Choice on Pretraining}

\begin{table}[t]
\caption{\textbf{Impact of visual encoder choice on pretraining.} Success rate on Move Playingcard Away and Click Bell for five model variants. Bridge variants are pretrained on Bridge Dataset v2; the OXE variant is pretrained on the large-scale Open X-Embodiment dataset.}
\centering
\footnotesize
\renewcommand{\arraystretch}{1.2}
\setlength{\tabcolsep}{3pt}
\newcommand{\taskcolwidth}{1.8cm}
\begin{tabular}{l|>{\centering\arraybackslash}p{\taskcolwidth}>{\centering\arraybackslash}p{\taskcolwidth}|>{\centering\arraybackslash}p{\taskcolwidth}>{\centering\arraybackslash}p{\taskcolwidth}}
\toprule
\textbf{Method}
  & \multicolumn{2}{c|}{\shortstack{Move Playingcard\\Away}}
  & \multicolumn{2}{c}{\shortstack{Click\\Bell}} \\
\cmidrule(lr){2-3}\cmidrule(lr){4-5}
& Easy & Hard & Easy & Hard \\
\midrule
OpenVLA-OFT + Frozen DINOv2 (Bridge)  & 0.04 & 0.08 & 0.18 & 0.12 \\
OpenVLA-OFT (Bridge)                   & 0.14 & 0.08 & 0.25 & 0.10 \\
OpenVLA-OFT + Frozen FiT3D (Bridge)   & \cellcolor[HTML]{FFF2CC}0.62 & 0.13 & \cellcolor[HTML]{E2EFDA}0.35 & \cellcolor[HTML]{E2EFDA}0.22 \\
OpenVLA-OFT + FiT3D (Bridge)          & 0.57 & \cellcolor[HTML]{FFF2CC}0.16 & 0.25 & 0.04 \\
\midrule
OpenVLA-OFT (OXE)                      & \cellcolor[HTML]{E2EFDA}0.70 & \cellcolor[HTML]{E2EFDA}0.34 & \cellcolor[HTML]{FFF2CC}0.20 & \cellcolor[HTML]{FFF2CC}0.08 \\
\bottomrule
\end{tabular}
\label{tab:encoder_pretrain}
\end{table}

Tab.~\ref{tab:encoder_pretrain} presents a controlled pretraining study comparing five model variants across two tasks of varying difficulty: Move Playingcard Away and Click Bell. All Bridge-pretrained variants are trained on Bridge Dataset v2~\citep{walke2023bridgedata} and evaluated on RoboTwin 2.0 after fine-tuning.

Several observations emerge from the results. First, freezing the original DINOv2 encoder yields the worst performance across both tasks, confirming that task-specific visual adaptation is necessary. Second, directly replacing DINOv2 with a frozen FiT3D encoder achieves substantially higher performance than both DINOv2 variants, demonstrating the strong intrinsic spatial quality of FiT3D features. Notably, on the relatively simple Move Playingcard Away task, the FiT3D Bridge-pretrained variant (0.62) approaches the performance of OpenVLA-OFT trained on the large-scale OXE dataset (0.70), despite using only a fraction of the training data. On the more spatially demanding Click Bell task, the frozen FiT3D variant (Easy: 0.35, Hard: 0.22) even surpasses the OXE-pretrained baseline (Easy: 0.20, Hard: 0.08), suggesting that richer spatial encoder features can partially compensate for limited pretraining data scale. These findings further motivate VEGA's design of transferring FiT3D spatial awareness into an existing VLA model via alignment, avoiding the prohibitive cost of full retraining from scratch.

\subsection{Simulation Visualization}
Fig.~\ref{fig:sim_vis} presents qualitative visualizations of VEGA executing all six RoboTwin 2.0 manipulation tasks. For each task, we display a sequence of keyframes illustrating the progression from initial state to task completion, demonstrating VEGA's ability to perform precise and spatially-aware manipulation across tasks of varying complexity.

\section{Real World Demostration}
We provide a supplementary video demonstrating VEGA on four real-world bimanual manipulation tasks: \textit{Close Laptop}, \textit{Handover Cucumber}, \textit{Pick Dual Carrots into Dual Bowls}, and \textit{Pick Dual Flowers into Vase}. The video showcases the policy execution across tasks of varying complexity, complementing the quantitative results reported in the main paper.

\section{Failure Case Analysis}

\begin{figure}[!t]
  \centering
  \includegraphics[width=\textwidth]{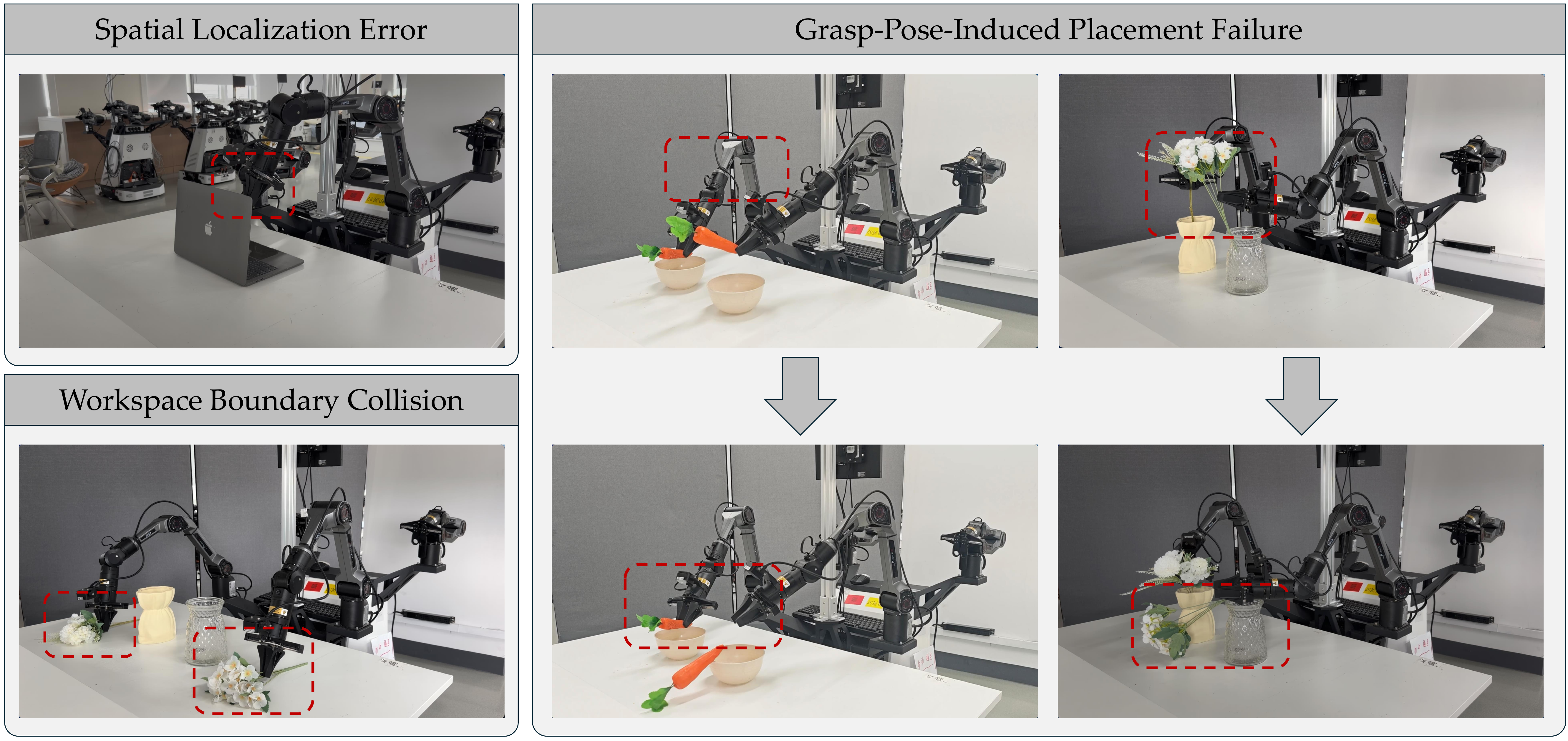}
   \caption{\textbf{Representative failure cases in real-world manipulation tasks.} (a) Spatial localization error. (b) Grasp-pose-induced placement failure. (c) Workspace boundary collision.}
  \label{fig:sim_vis}
\end{figure}

To better understand the limitations of VEGA in real-world settings, we analyze representative failure cases observed across the four manipulation tasks, grouping them into three categories based on their underlying failure mechanism, as illustrated in Fig.~\ref{fig:sim_vis}. 

\paragraph{Type I: Spatial Localization Error.} In the \textit{Close Laptop} task, the gripper occasionally approaches the screen side of the laptop rather than the correct rear surface near the hinge. Once committed to this trajectory, the policy fails to self-correct, and the episode terminates upon reaching the maximum step limit. This failure reflects residual imprecision in 3D spatial localization of articulated object parts, suggesting that while VEGA substantially improves spatial grounding, fine-grained part-level localization of thin or occluded structures remains challenging.

\paragraph{Type II: Grasp-Pose-Induced Placement Failure.} Two tasks share a common failure mode in which a suboptimal grasp pose leads to downstream placement failure. In \textit{Pick Dual Carrots into Dual Bowls}, when the gripper grasps the carrot near its base rather than its center, the effective length of the held object exceeds the bowl's spatial tolerance, causing the carrot to miss or fall out during insertion. Similarly, in \textit{Pick Dual Flowers into Vase}, an off-center grasp on the flower stem results in a non-vertical object orientation, preventing successful insertion into the narrow vase opening. In both cases, the placement error is not a failure of spatial perception per se, but rather a consequence of grasp pose variance propagating into the placement stage—a limitation inherent to open-loop execution without explicit grasp quality estimation.

\paragraph{Type III: Workspace Boundary Collision.} Also, in \textit{Pick Dual Flowers into Vase}, the thin flower stems lie in close proximity to the tabletop surface. During the grasping motion, the robot arm occasionally makes contact with the table, triggering a safety-induced disabling of the actuators and causing task failure. This failure mode is primarily attributable to insufficient clearance estimation near the workspace boundary, and is exacerbated by the visual ambiguity of thin, low-contrast objects against the tabletop.

\section{Broader Impacts}

This work proposes VEGA, a visual encoder alignment framework aimed at improving spatial perception in Vision-Language-Action (VLA) models for robot manipulation. On the positive side, advances in robot spatial understanding can benefit a wide range of applications, including assistive robotics for individuals with physical disabilities, automation of repetitive or hazardous tasks, and more reliable human-robot collaboration in manufacturing and healthcare settings.

On the negative side, improved robot manipulation capabilities may contribute to concerns around labor displacement in industries that rely on manual dexterity. Furthermore, as VLA models become more capable, ensuring safe and predictable behavior in unstructured real-world environments remains an open challenge; incorrect spatial reasoning could lead to unintended physical consequences in human-adjacent deployments. We believe that responsible deployment practices, including thorough safety evaluation and staged real-world testing, are essential to mitigate these risks. This work does not involve human subjects, personal data, or applications with direct dual-use concerns.

\end{document}